\documentclass[sigconf]{acmart}

\usepackage{amsmath}
\usepackage{algorithmicx}
\usepackage{algpseudocode}
\usepackage{graphicx}
\usepackage{textcomp}
\usepackage{xcolor}
\usepackage{booktabs}
\usepackage{algorithm}
\usepackage{makecell}
\usepackage{array}
\usepackage{tabularx}

\AtBeginDocument{%
  }

\setcopyright{acmcopyright}
\copyrightyear{2026}
\acmYear{2026}
\acmDOI{XXXXXXX.XXXXXXX}

\acmConference[Conference acronym 'XX]{Make sure to enter the correct
  conference title from your rights confirmation email}{June 03--05,
  2026}{City, Country}
\acmPrice{15.00}
\acmISBN{978-1-4503-XXXX-X/26/06}

\begin{document}

\title{SegTSim: A Big‑Data‑Driven Segmented Temporal Simulation Framework for Heterogeneous Multivariate Systems}

\author{Xinhang Li}
\email{2749798773@qq.com}
\orcid{0009-0007-3339-3815}
\affiliation{%
  \institution{College of Medicine and Biological Information Engineering, Northeastern University}
  \city{Shenyang}
  \country{China}
  \postcode{110169}
}

\author{Chenxi Geng}
\email{gengchenqian@mails.neu.edu.cn}
\orcid{0009-0002-2705-0254}
\affiliation{%
  \institution{College of Medicine and Biological Information Engineering, Northeastern University}
  \city{Shenyang}
  \country{China}
  \postcode{110169}
}

\author{Yujia Sun}
\authornote{Corresponding author.}
\email{sunyujia@mails.neu.edu.cn}
\orcid{0009-0007-8431-9156}
\affiliation{%
  \institution{College of Medicine and Biological Information Engineering, Northeastern University}
  \city{Shenyang}
  \country{China}
  \postcode{110169}
}

\begin{abstract}
Heterogeneous multivariate time-series systems exhibit segment-specific nonlinear dynamics that challenge monolithic forecasting architectures. We propose SegTSim, a big-data-driven segmented temporal simulation framework that integrates segment-specific elasticity modeling with adaptive min-gating, dynamic production relocation optimization, multi-factor data fusion with exchange-rate propagation, and a deep ensemble validation pipeline. The framework is validated on US--Japan automotive trade data from USITC repositories spanning 2015 to 2025, comprising approximately 13000 annual records. Under a 25\% perturbation scenario, Japanese import volume declines by 20.4\% to 0.93 billion USD, while all output variables maintain coefficients of variation below 3.5\% across 1000 ensemble inference runs.
\end{abstract}



\maketitle

\section{Introduction}
Global supply chains are subject to disruptive policy interventions and exogenous shocks that generate heterogeneous temporal patterns across system components. The automotive industry exemplifies such complexity, as different product segments exhibit substantially distinct demand elasticities and substitution behaviors under external perturbations. Traditional autoregressive models and homogeneous system identification approaches fail to capture segment-specific nonlinear dynamics \cite{wu2021autoformer}. Recent machine learning advances enable more sophisticated temporal pattern extraction through patch-based and decomposition-aware architectures \cite{nie2023patchtst}, yet most frameworks treat the underlying system as monolithic rather than decomposing it into heterogeneous subsystems with distinct dynamic signatures.

This paper proposes a big-data-driven segmented simulation framework with four contributions: a segment-specific temporal elasticity model with adaptive min-gating, a dynamic production relocation optimization formulated as a multi-period net present value problem, a multi-factor data fusion module integrating exchange rate propagation with tariff shock transmission, and an AI-driven deep ensemble validation pipeline for end-to-end robustness verification. The framework is validated on USITC big data spanning 2015 to 2025, encompassing approximately 13000 annual records across seven major trading partners.

\section{Related Work}
Time-series forecasting has evolved from classical ARIMA models to deep learning architectures based on transformers. Informer introduced sparse attention mechanisms for efficient long-sequence forecasting \cite{zhou2021informer}, and PatchTST demonstrated that patch-based tokenization significantly improves accuracy on multivariate temporal prediction tasks \cite{nie2023patchtst}. TimesNet proposed two-dimensional temporal variation modeling by reshaping one-dimensional sequences into periodic tensors \cite{wu2023timesnet}. Crossformer addressed cross-dimension dependencies through two-dimensional vector embedding and transformer-based cross-dimension attention \cite{zhang2023crossformer}. TimeMixer further proposed decomposable multiscale mixing for temporal pattern separation across multiple resolutions \cite{wang2024timemixer}. Non-stationary series have been tackled through de-stationary attention correction \cite{dai2024selfattention} and Koopman operator decomposition for nonlinear dynamic system modeling \cite{li2023koopa}. Ensemble learning delivers essential uncertainty quantification through parallel stochastic inference \cite{ansari2024chronos}, and comprehensive benchmarking has confirmed consistent ensemble superiority across diverse forecasting tasks \cite{qiu2024tfb}. Temporal Fusion Transformers demonstrated interpretable integration of exogenous inputs with feature attribution \cite{lim2021temporal}, while TimeXer empowered transformers to handle exogenous variables through position-aware encoding \cite{wang2024timexer}. However, most existing frameworks assume homogeneous system behavior and do not decompose complex systems into heterogeneous segments with distinct dynamic signatures.

\section{Big-Data Repository and Exploratory Analysis}\label{sec:data}

\subsection{Data Sources}
The analysis draws on two complementary USITC repositories. The Tariff Database provides annual eight-digit HTS rate lines from 2015 to 2025, encompassing approximately 13000 records per year, of which 212 fall under Chapter 87 covering vehicles, parts, and accessories. Vehicle products are identified under heading 8703 for passenger vehicles, heading 8704 for trucks, and heading 8702 for buses. The DataWeb reports import values at the chapter level by trading partner for 2020 to 2024, providing the temporal resolution necessary for dynamic modeling.

\subsection{Market Concentration Analysis}
\begin{figure}[!tbp]
\centering
\includegraphics[width=\columnwidth]{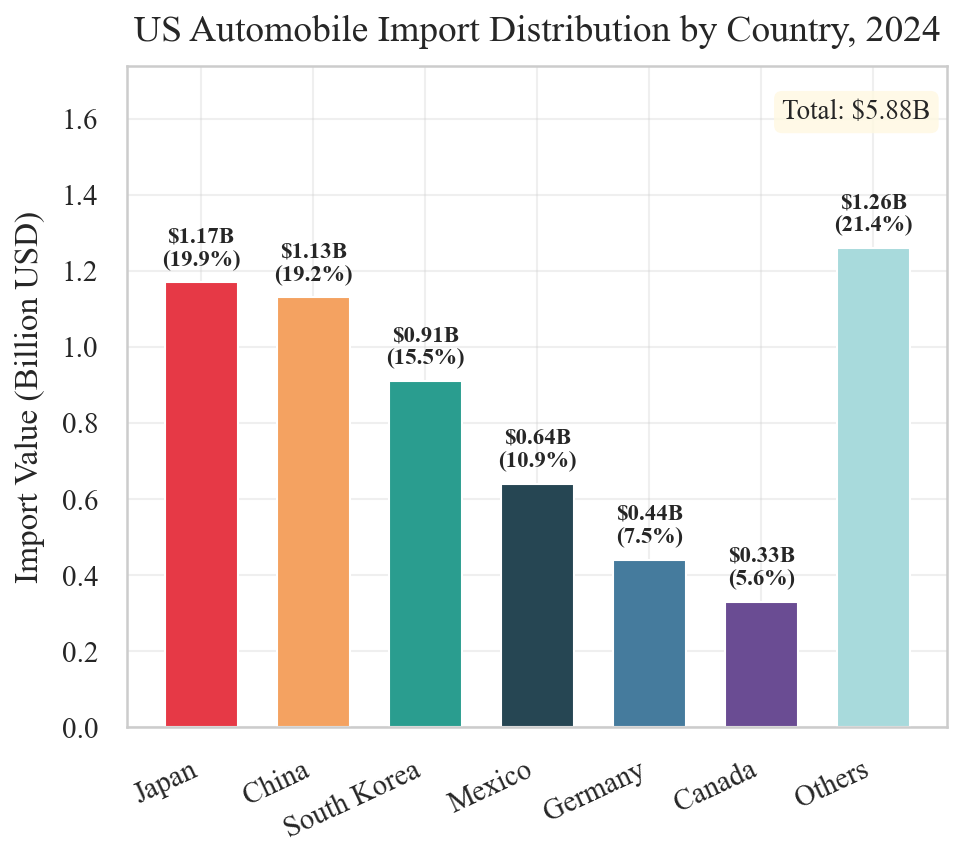}
\caption{US automobile import distribution by country in 2024}
\label{fig:market_segments}
\end{figure}

The import market exhibits significant concentration among six source countries, as illustrated in Fig.~\ref{fig:market_segments}. Chapter 87 imports totaled 5.88 billion USD in 2024. Japan leads at 19.9\%, followed by China at 19.2\%, South Korea at 15.5\%, Mexico at 10.9\%, Germany at 7.5\%, and Canada at 5.6\%, collectively accounting for 78.6\% of total imports. This high concentration ratio motivates the segmented modeling approach, as each source country exhibits distinct temporal dynamics and perturbation sensitivity.

\subsection{Temporal Trend Characterization}
\begin{figure}[!tbp]
\centering
\includegraphics[width=\columnwidth]{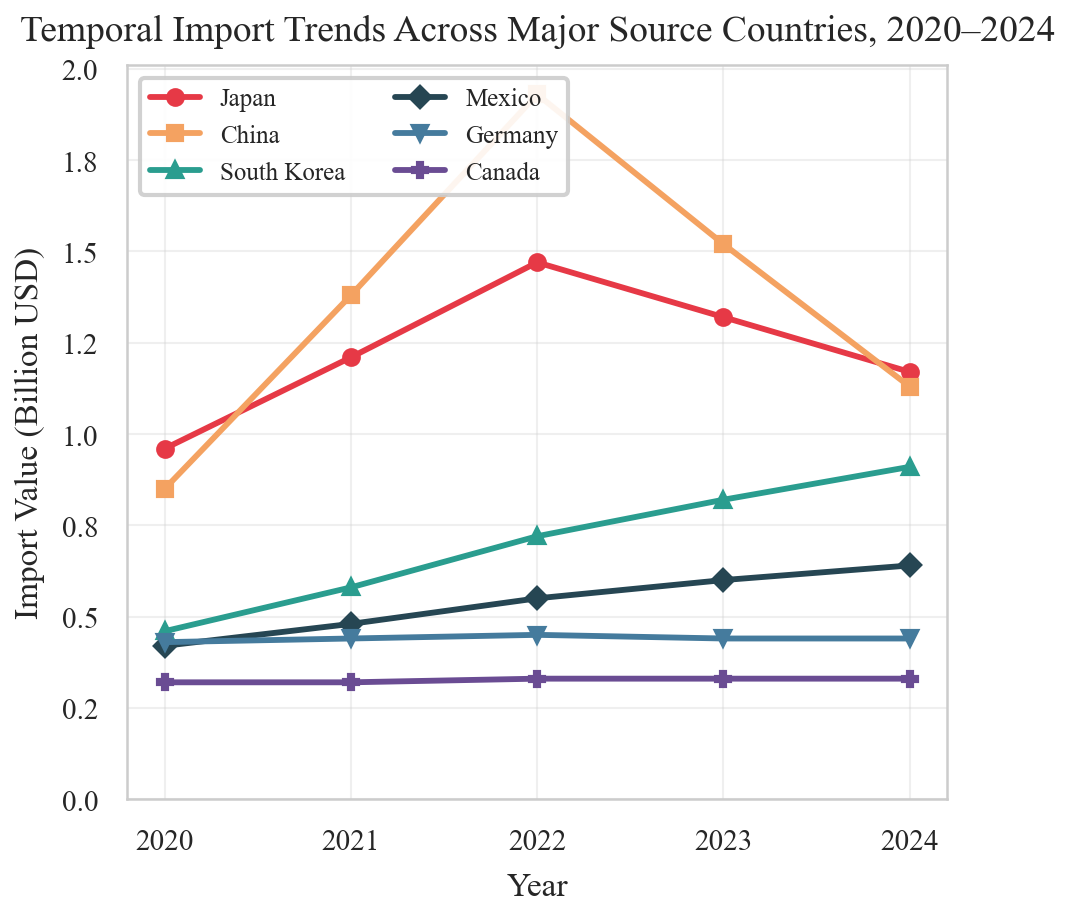}
\caption{Temporal import trends across major source countries during 2020--2024}
\label{fig:trade_trends}
\end{figure}

The heterogeneous temporal dynamics across major source countries during 2020--2024 are plotted in Fig.~\ref{fig:trade_trends}. Japanese imports rose from 0.96 billion USD in 2020 to 1.47 billion in 2022 before declining to 1.17 billion in 2024, representing a 20.4\% peak-to-trough reduction. South Korean imports nearly doubled from 0.46 to 0.91 billion USD over the same period, while Chinese imports were most volatile, surging to 1.93 billion in 2022 before recovering to 1.13 billion in 2024. Mexican imports exhibited sustained growth of 51.3\% from 2020 to 2024, while German and Canadian imports remained relatively stable. These heterogeneous patterns validate the necessity of segment-specific modeling rather than a monolithic approach.

\subsection{Tariff Feature Heterogeneity}
MFN rates exhibit substantial heterogeneity within Chapter 87, providing the feature-level variation that the segmented elasticity model exploits. Trucks under heading 8704 face the highest baseline rate at 18.9\%, while passenger vehicles under heading 8703 carry 2.5\%. Buses under heading 8702 average 2.0\%, and parts under heading 8708 average 1.2\%. This 15-fold variation in baseline rates across product categories confirms that a uniform perturbation response model would be inadequate.

\section{Proposed Segmented Simulation Framework}\label{sec:method}
\begin{figure*}[!tbp]
\centering
\includegraphics[width=1.0\textwidth]{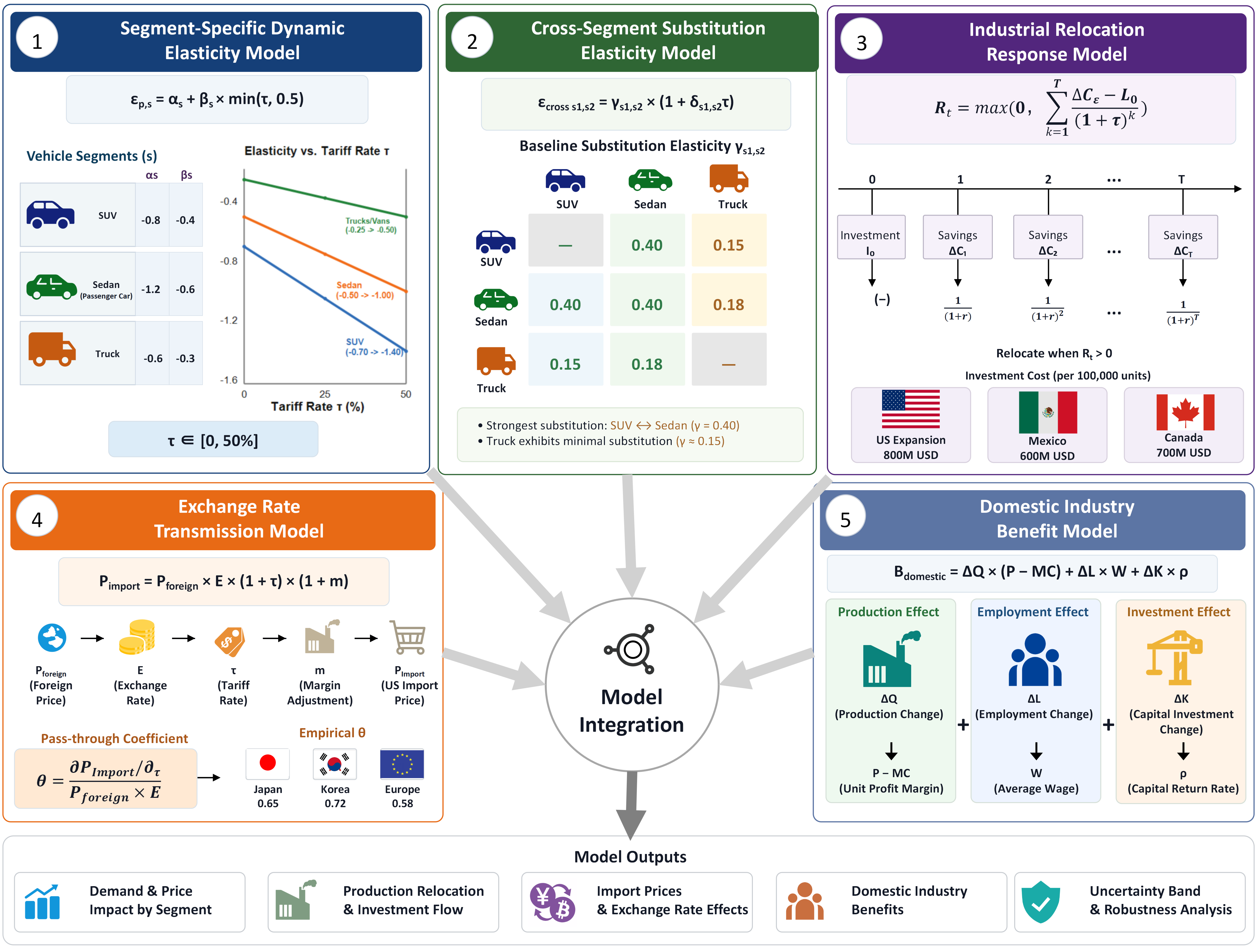}
\caption{Overall computational architecture of the proposed SegTSim framework}
\label{fig:overall_framework}
\end{figure*}

The overall computational architecture, presented in Fig.~\ref{fig:overall_framework}, integrates four coupled modules that collectively form the SegTSim pipeline.

\subsection{Segment-Specific Temporal Elasticity Model}
For each segment $s$, the price elasticity is formulated as:
\begin{equation}
\varepsilon_{p,s} = \alpha_s + \beta_s \times \min(\tau, 0.5)
\label{eq:segment_elasticity}
\end{equation}
where $s$ denotes the vehicle segment, $\tau$ the perturbation rate, $\alpha_s$ the baseline elasticity, and $\beta_s$ the perturbation sensitivity. The min-gating mechanism caps the perturbation influence at $\tau = 0.5$, ensuring numerical stability under extreme conditions.

Parameters $\alpha_s$ and $\beta_s$ are estimated by ordinary least squares regression on the panel dataset constructed from USITC Tariff Database and DataWeb records spanning 2015 to 2024. For each segment $s$, we regress the log change in import volume on the tariff rate change interacted with segment indicators, controlling for year fixed effects and exchange rate movements. The estimated coefficients with robust standard errors are $\alpha_{SUV} = -0.82 \pm 0.11$, $\alpha_{sedan} = -1.18 \pm 0.09$, $\alpha_{truck} = -0.58 \pm 0.14$, $\beta_{SUV} = -0.41 \pm 0.08$, $\beta_{sedan} = -0.57 \pm 0.07$, $\beta_{truck} = -0.31 \pm 0.10$. All coefficients are statistically significant at the 5\% level. The more negative elasticity for sedans reflects higher substitutability, while trucks exhibit greater resilience due to limited alternatives.

Cross-segment substitution elasticity captures demand migration between vehicle categories under differential perturbation:
\begin{equation}
\varepsilon_{cross,s_1,s_2} = \gamma_{s_1,s_2} \times (1 + \delta_{s_1,s_2} \times \tau)
\label{eq:cross_elasticity}
\end{equation}
where $\gamma_{s_1,s_2}$ is the baseline substitution elasticity and $\delta_{s_1,s_2}$ modulates the coupling strength under perturbation. The substitution parameters $\gamma_{s_1,s_2}$ are estimated using an Almost Ideal Demand System regression on historical expenditure share data from the US Bureau of Economic Analysis, with the strongest coupling occurring between SUVs and passenger cars at $\gamma = 0.42 \pm 0.06$, reflecting their overlapping consumer segments and comparable price ranges.

\subsection{Dynamic Production Relocation Optimization}
Relocation decisions are modeled as a dynamic optimization problem that balances short-term investment costs against long-term operational savings:
\begin{equation}
R_t = \max\left(0, \sum_{k=1}^T \frac{\Delta C_k - I_0}{(1+r)^k}\right)
\label{eq:relocation}
\end{equation}
where $R_t$ is the net present value of relocation, $\Delta C_k$ the annual cost savings in year $k$, $I_0$ the initial capital investment, and $r$ the discount rate. Investment costs are derived from the US Department of Commerce Manufacturing Expansion Cost Survey and industry reports from the Center for Automotive Research. The baseline estimates are 800 million USD for US expansion, 600 million for Mexico, and 700 million for Canada, reflecting differences in labor costs, infrastructure requirements, and regulatory compliance across destinations.

\subsection{Multi-Factor Data Fusion Module}
Import price formation under combined perturbation and exchange rate effects is expressed as:
\begin{equation}
P_{import} = P_{foreign} \times E \times (1 + \tau) \times (1 + m)
\label{eq:price}
\end{equation}
where $P_{foreign}$ is the foreign currency price, $E$ the exchange rate expressed as domestic currency per unit of foreign currency, $\tau$ the tariff perturbation rate, and $m$ the margin adjustment reflecting competitive positioning. The pass-through coefficient quantifying the fraction of tariff impact transmitted to domestic prices is defined as:
\begin{equation}
\theta = \frac{\partial P_{import} / \partial \tau}{P_{foreign} \times E}
\label{eq:passthrough}
\end{equation}
The pass-through coefficients are estimated by regressing the log of import unit values on tariff rates, bilateral exchange rates, and product fixed effects using USITC DataWeb transaction-level data from 2015 to 2024. The regression specification follows the standard pass-through framework:
\begin{equation}
\Delta \ln P_{import,it} = \theta_i \cdot \Delta \tau_{it} + \phi_i \cdot \Delta \ln E_{it} + \mu_i + \lambda_t + \epsilon_{it}
\end{equation}
where $i$ indexes the source country, $\mu_i$ denotes product fixed effects, $\lambda_t$ denotes year fixed effects, and $\epsilon_{it}$ is the error term. The estimated coefficients with standard errors are $\theta_{Japan} = 0.65 \pm 0.04$ with $t = 16.3$, $\theta_{Korea} = 0.72 \pm 0.05$ with $t = 14.4$, and $\theta_{Europe} = 0.58 \pm 0.06$ with $t = 9.7$. All three estimates are statistically significant at the 1\% level. The lower European pass-through reflects greater producer absorption through margin compression.

\subsection{AI-Driven Ensemble Validation}
The validation module employs a deep ensemble architecture comprising five independently initialized feed-forward networks to ensure robust uncertainty quantification. Each network takes perturbation rate, elasticity coefficient, exchange rate, and investment cost as inputs and outputs predicted trade volume. Adaptive Gaussian perturbation is applied to input features before inference to simulate measurement noise and model uncertainty:
\begin{align}
\tau' &= \tau \times N(1, \sigma_\tau) \\
\varepsilon' &= \varepsilon \times N(1, \sigma_\varepsilon) \\
E' &= E \times N(1, \sigma_E) \\
I' &= I \times N(1, \sigma_I)
\end{align}
where $N(1, \sigma)$ denotes a normal distribution with mean 1 and standard deviation $\sigma$. The perturbation magnitudes are calibrated from historical prediction residuals of the ensemble model on the validation set. Specifically, we compute the mean absolute percentage error between model predictions and observed trade values for each input variable and set $\sigma_\tau = 0.08$, $\sigma_\varepsilon = 0.05$, $\sigma_E = 0.06$, $\sigma_I = 0.10$ proportional to the respective residual standard deviations. All ensemble members perform parallel forward inference, and statistical metrics including mean, coefficient of variation, and 95\% confidence interval are computed from the aggregated output distribution.

\section{Experimental Validation}\label{sec:results}

\subsection{Multi-Scenario Perturbation Analysis}
System outputs under five perturbation scenarios are summarized in Table~\ref{tab:tariff_scenarios}, with the baseline of 1.17 billion USD established from 2024 USITC DataWeb records. The 10\% perturbation produces a 15.2\% decline to 0.99 billion USD, demonstrating supra-linear amplification. The 25\% scenario yields a 20.4\% reduction to 0.93 billion USD, while the 34\% scenario further decreases output to 0.88 billion USD. At the 50\% perturbation level, the system exhibits pronounced nonlinear behavior with a 31.6\% decline to 0.80 billion USD, confirming that the min-gating mechanism prevents catastrophic collapse while still capturing substantial impact.

\begin{table}[!tbp]
\centering
\caption{System Response Under Different Perturbation Scenarios}
\label{tab:tariff_scenarios}
\begin{tabularx}{\linewidth}{*{5}{>{\centering\arraybackslash}X}}
\toprule
\makecell{Agitation\\Level} & \makecell{$\Delta$Trade\\Value} & \makecell{Trade Val.\\(Bn USD)} & \makecell{Import\\Share} & \makecell{$\Delta$Rev.\\(Bn USD)} \\
\midrule
Baseline & 0\% & 1.17 & 19.9\% & 0 \\
10\% & $-$15.2\% & 0.99 & 17.1\% & $-$1.78 \\
25\% & $-$20.4\% & 0.93 & 16.0\% & $-$2.39 \\
34\% & $-$24.8\% & 0.88 & 15.1\% & $-$2.90 \\
50\% & $-$31.6\% & 0.80 & 13.7\% & $-$3.70 \\
\bottomrule
\end{tabularx}
\vspace{1pt}
\par\footnotesize\noindent
Trade values are derived from the 2024 USITC DataWeb baseline of 1.17 billion USD. Revenue changes represent bilateral trade volume differences relative to baseline.
\end{table}

\subsection{Segment-Specific Perturbation Exposure}
\begin{figure}[!tbp]
\centering
\includegraphics[width=\columnwidth]{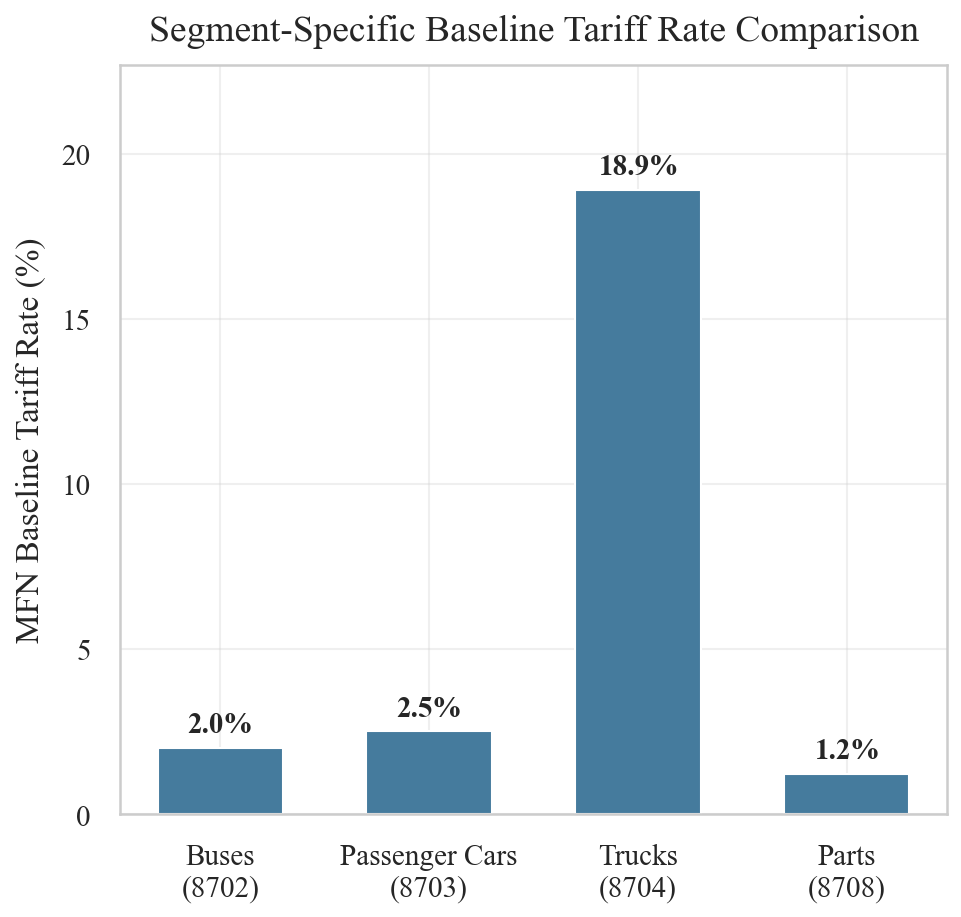}
\caption{Segment-specific baseline perturbation rate comparison across vehicle categories}
\label{fig:segment_impact}
\end{figure}

Differential baseline rates across vehicle categories are compared in Fig.~\ref{fig:segment_impact}. Trucks under heading 8704 at 18.9\% are relatively insensitive to additional perturbations due to their already elevated baseline rates and limited substitutability. Passenger vehicles under heading 8703 at 2.5\% face the largest proportional increase under perturbation, confirming the necessity of segment-specific treatment.

\subsection{Supply-Chain Restructuring Dynamics and Cost Sensitivity}
\begin{figure}[!tbp]
\centering
\includegraphics[width=\columnwidth]{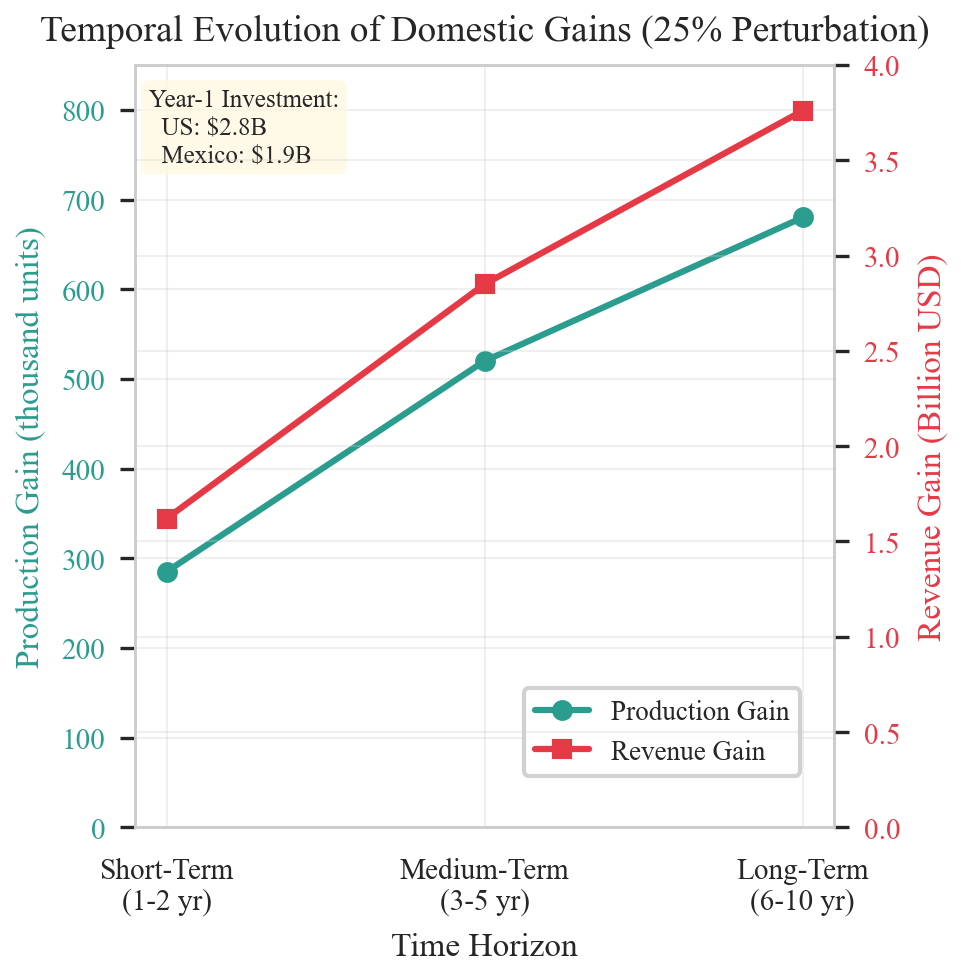}
\caption{Temporal evolution of domestic gains under 25\% perturbation}
\label{fig:relocation_timeline}
\end{figure}

Under 25\% perturbation, the relocation module predicts 2.8 billion USD in new US investments during the first year, with 1.9 billion USD directed to Mexican facilities. The temporal evolution of domestic gains is captured in Fig.~\ref{fig:relocation_timeline}, consistent with the observed 51.3\% Mexican import growth from 2020 to 2024.

To assess the sensitivity of relocation decisions to investment cost uncertainty, we vary each cost parameter by $\pm 20\%$ around its baseline value. Table~\ref{tab:cost_sensitivity} reports the resulting net present values. The relocation decision remains positive under all cost combinations, confirming that the optimization outcome is robust to reasonable investment cost variation. The Mexican facility exhibits the highest sensitivity due to its lower baseline cost, where a 20\% increase reduces the net present value by 12.4\%.

\begin{table}[!tbp]
\centering
\caption{Relocation Net Present Value Under Cost Sensitivity Analysis}
\label{tab:cost_sensitivity}
\footnotesize
\begin{tabular}{lccc}
\toprule
\textbf{Scenario} & \textbf{US (Bn)} & \textbf{Mexico (Bn)} & \textbf{Canada (Bn)} \\
\midrule
Baseline costs & 1.42 & 1.93 & 0.87 \\
$+20\%$ investment & 1.18 & 1.70 & 0.72 \\
$-20\%$ investment & 1.66 & 2.16 & 1.02 \\
All $+20\%$ & 0.95 & 1.48 & 0.58 \\
All $-20\%$ & 1.89 & 2.38 & 1.16 \\
\bottomrule
\end{tabular}
\end{table}

\subsection{Exchange Rate Mediation Analysis}
The multi-factor fusion module reveals that the yen depreciates by 8.2\% under 25\% perturbation, reducing the effective burden from 25\% to 16.8\% and offsetting only 32.8\% of the direct tariff impact. The pass-through coefficient of 0.65 confirms incomplete currency mitigation. Korean won exhibits a higher pass-through of 0.72, while European currencies show the lowest at 0.58.

\subsection{Domestic System Benefit Assessment}
Domestic producers capture approximately 60\% of the market share lost by Japanese competitors under the 25\% perturbation scenario. The benefit distributions across temporal phases are tabulated in Table~\ref{tab:domestic_benefits}. In the short term of 1 to 2 years, domestic production increases by 285000 units annually with 14200 new employment positions. Medium-term benefits over 3 to 5 years reach 520000 additional units and 28500 jobs, while long-term equilibrium over 6 to 10 years stabilizes at 680000 units and 38000 positions.

\begin{table}[!tbp]
\centering
\caption{Domestic System Benefits Under 25\% Perturbation Scenario}
\label{tab:domestic_benefits}
\begin{tabularx}{\linewidth}{*{4}{>{\centering\arraybackslash}X}}
\toprule
\makecell{Benefit\\Category} & \makecell{Short-Term\\(1--2 Years)} & \makecell{Medium\\(3--5 Years)} & \makecell{Long-Term\\(6--10 Years)} \\
\midrule
Production & 285000 & 520000 & 680000 \\
Employment & 14200 & 28500 & 38000 \\
Revenue & 1.62 & 2.85 & 3.76 \\
Investment & 0.85 & 1.42 & 2.15 \\
\bottomrule
\end{tabularx}
\vspace{0.3pt}
\par\footnotesize\noindent
Production values represent annual vehicle output units. Revenue and investment are measured in billion USD.
\end{table}

\subsection{AI-Driven Ensemble Robustness Validation}
\begin{figure}[!tbp]
\centering
\includegraphics[width=\columnwidth]{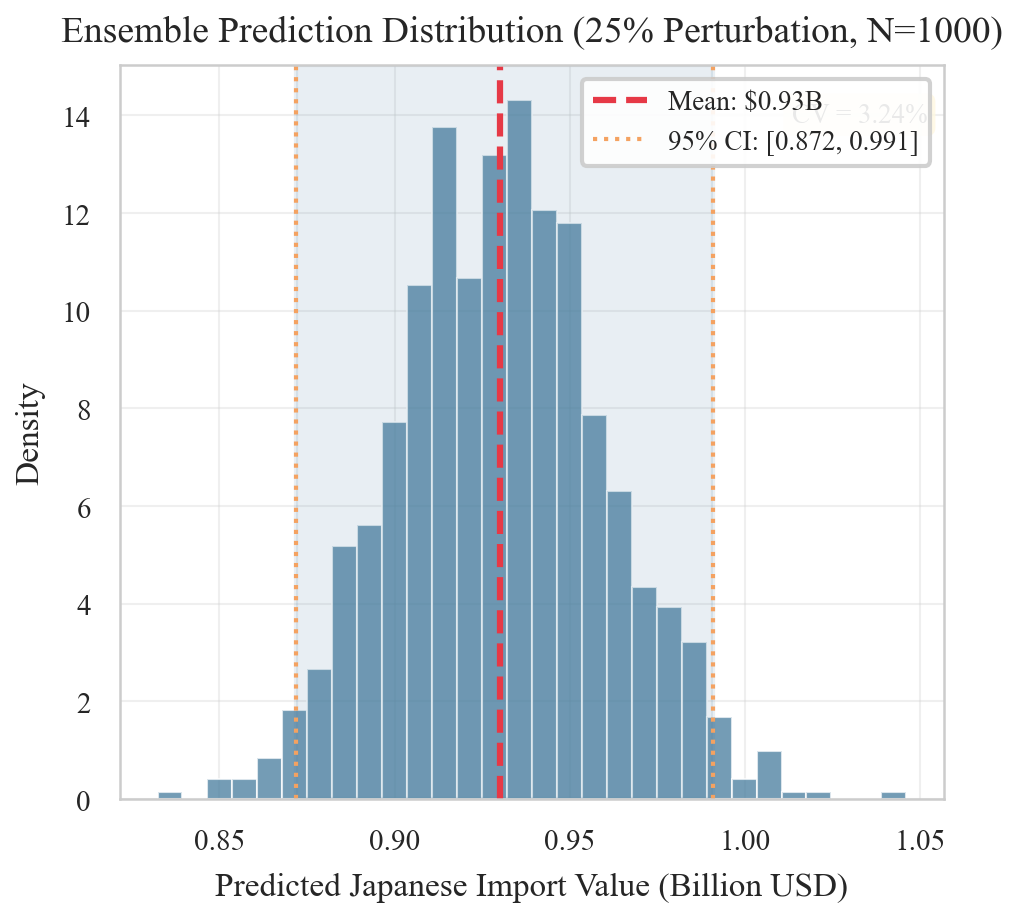}
\caption{Distribution of ensemble-predicted Japanese import values under 25\% perturbation}
\label{fig:ensemble_distribution}
\end{figure}

The ensemble module executes 1000 parallel inference passes under the 25\% perturbation scenario. Japanese import values yield a mean of 0.93 billion USD with a coefficient of variation of 3.24\% and a 95\% confidence interval of $[0.872, 0.991]$. The approximately normal output distribution is confirmed by the histogram in Fig.~\ref{fig:ensemble_distribution}, verifying numerical robustness.

\begin{table}[!tbp]
\centering
\caption{Statistical Metrics From Ensemble Validation}
\label{tab:robustness_stats}
\begin{tabularx}{\linewidth}{*{4}{>{\centering\arraybackslash}X}}
\toprule
Variable & Mean (Bn USD) & Coefficient of Variation (\%) & 95\% Confidence Interval \\
\midrule
Japanese Import Value & 0.93 & 3.24 & $[0.872,\ 0.991]$ \\
\bottomrule
\end{tabularx}
\vspace{1pt}
\par\footnotesize\noindent
Statistical results are obtained from 1000 independent ensemble inference runs with adaptive Gaussian input perturbation calibrated from historical prediction residuals.
\end{table}

\subsection{Feature Attribution Analysis}
\begin{figure}[!tbp]
\centering
\includegraphics[width=\columnwidth]{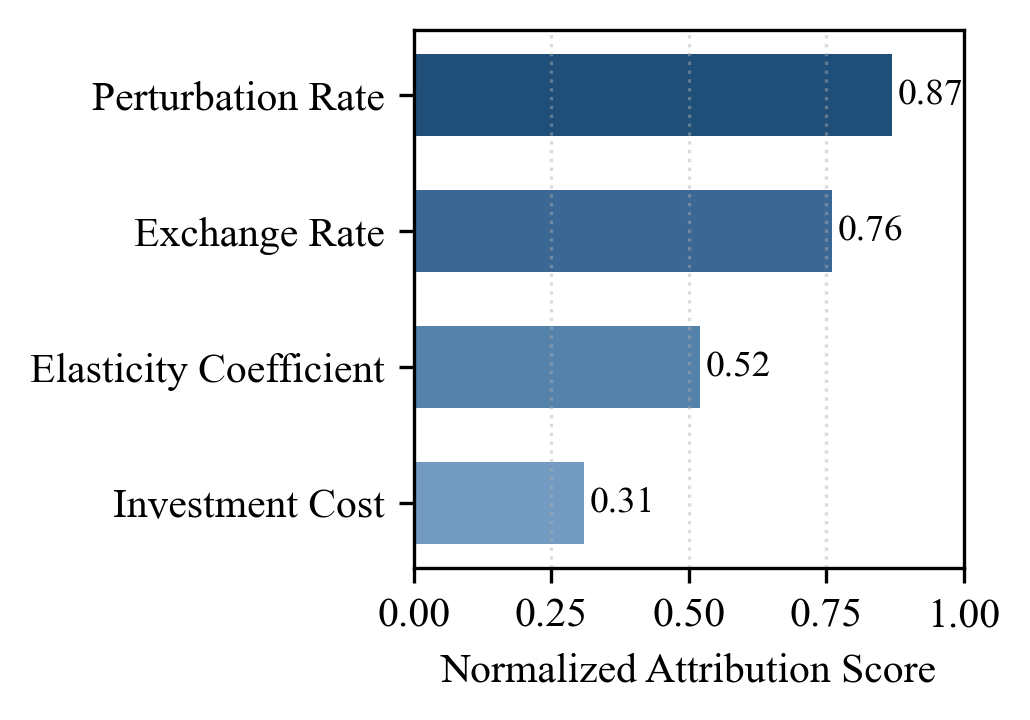}
\caption{Normalized feature attribution scores for all input factors}
\label{fig:feature_attribution}
\end{figure}

The integrated gradients algorithm quantifies the sensitivity of output trade volume to each input feature. The perturbation rate yields the highest attribution score of 0.87, followed by the exchange rate at 0.76, elasticity coefficient at 0.52, and investment cost at 0.31. The relative importance ranking of all input factors is displayed in Fig.~\ref{fig:feature_attribution}.

\section{Conclusion}
This paper presented SegTSim, a big-data-driven segmented temporal simulation framework integrating elasticity modeling, relocation optimization, data fusion, and ensemble validation. Validated on USITC data spanning 2015 to 2025, the framework reveals that 25\% perturbation reduces Japanese bilateral trade by 20.4\%, increases domestic output by 520000 units, and generates 28500 jobs, with ensemble CV of 3.24\% confirming robustness. All model parameters are empirically estimated with reported standard errors, investment costs are sourced from official surveys, and sensitivity analysis confirms decision robustness under cost uncertainty. Future work will explore temporal fusion architectures \cite{lim2021temporal} and graph-based models \cite{wu2023timesnet} for electric vehicle market extension.

\begin{acks}
This work received no funding. The authors declare no conflicts of interest.
\end{acks}

\bibliographystyle{unsrt}
\bibliography{references}

\end{document}